\documentclass{article}
\usepackage{spconf,amsmath,graphicx}
\usepackage{todonotes}
\usepackage{indentfirst}
\usepackage{url}
\usepackage{hyperref}
\usepackage{float}
\usepackage{svg}
\usepackage{array}
\usepackage{booktabs}
\usepackage{multirow}
\usepackage{subcaption}

\title{MMP-2k: A Benchmark Multi-labeled Macro Photography Image Quality Assessment Database }
\name{Jiashuo Chang$^{1,\dagger}$, Zhengyi Li$^{1,\dagger}$, Jianxun Lou$^2$, Zhen Qiu$^3$, Hanhe Lin$^{1,3}$ \thanks{$^\dagger$ Jiashuo Chang and Zhengyi Li contribute equally.
}}
\address{$^1$Dundee International Institute of Central South University, Central South University, China\\
$^2$School of Computer Science, Northeast Electric Power University, China \\
$^3$School of Science and Engineering, University of Dundee, United Kingdom\\
}
\begin{document}
%
\maketitle
\begin{abstract}

Macro photography (MP) is a specialized field of photography that captures objects at an extremely close range, revealing tiny details. Although an accurate macro photography image quality assessment (MPIQA) metric can benefit macro photograph capturing, which is vital in some domains such as scientific research and medical applications, the lack of MPIQA data limits the development of MPIQA metrics.  To address this limitation, we conducted a large-scale MPIQA study. Specifically, to ensure diversity both in content and quality, we sampled 2,000 MP images from 15,700 MP images, collected from three public image websites. For each MP image, 17 (out of 21 after outlier removal) quality ratings and a detailed quality report of distortion magnitudes, types, and positions are gathered by a lab study. 
The images, quality ratings, and quality reports form our novel multi-labeled MPIQA database, MMP-2k. Experimental results showed that the state-of-the-art generic IQA metrics underperform on MP images. 
The database and supplementary materials are available at \url{https://github.com/Future-IQA/MMP-2k}.

\end{abstract}
\begin{keywords}
Macro photography, image quality assessment, subjective study, benchmark database
\end{keywords}
\section{Introduction}
\label{sec:intro}
Macro photography (MP) is the extreme close-up photography that shows small objects at a large scale usually invisible to the naked eye \cite{cosentino2013macro}. Typically, a photograph of small objects with a magnification of ``life-size" or greater, short focal distances and a shallow depth of field is considered a macro photograph~\cite{adobe2024}. Macro photography is a powerful tool that has a wide range of applications. Shutterbugs use macro photography to capture intricate details in flowers and insects that show the beauty of the natural world. Clinical photographers take macro photographs to document skin conditions, surgical procedures, and tissue samples, to name a few. Despite the importance of MP, taking a high-quality macro photograph is challenging since it involves a few techniques different from shooting regular normal-distance photos. In light of this, developing an accurate macro photography image quality assessment (MPIQA) metric could significantly improve high-quality macro photograph capturing.

\begin{figure}[t]
        \centering
        \includegraphics[width=1.0\linewidth]{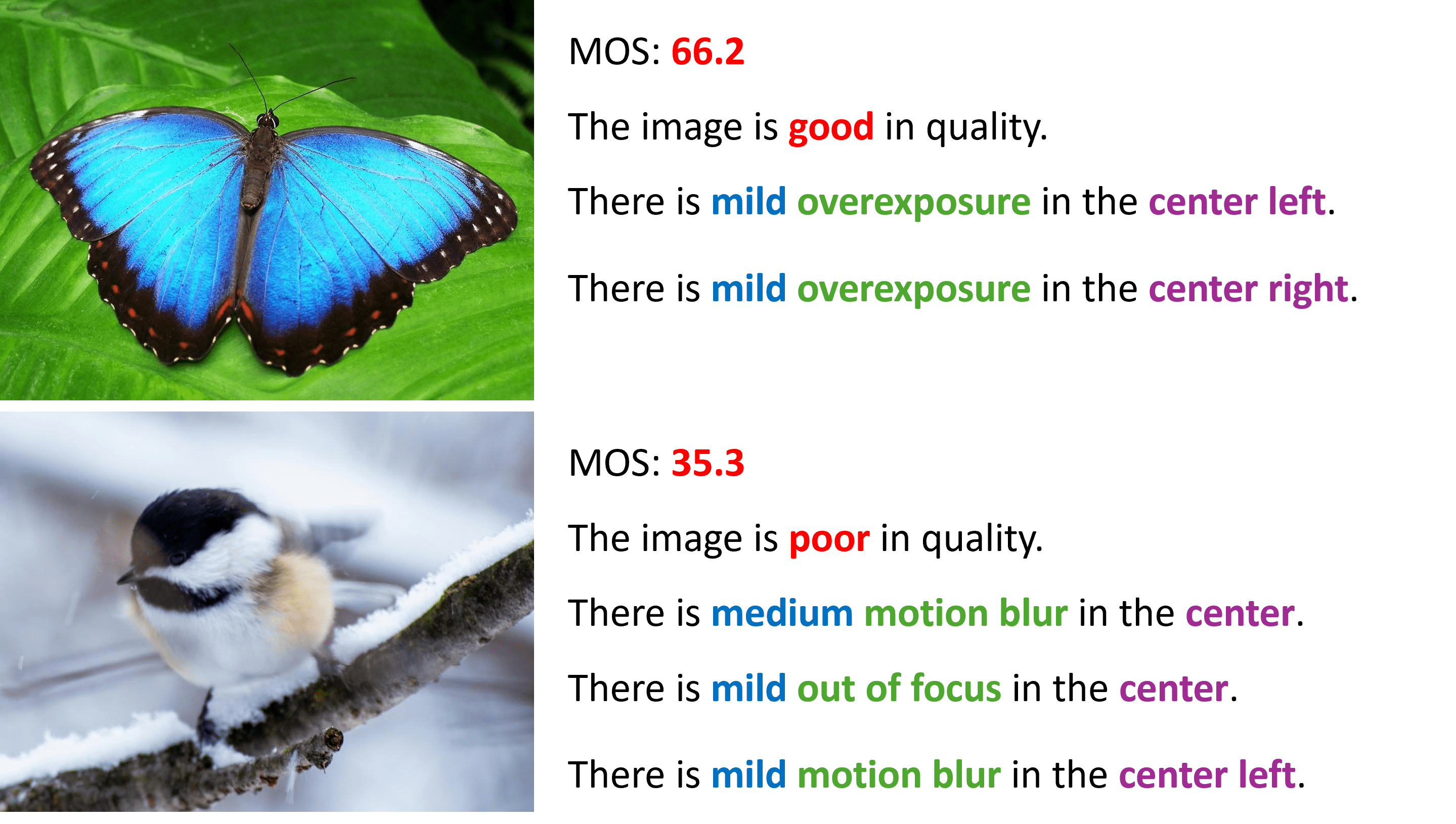}
        \caption{MP images in the MMP-2k database have two labels, a MOS (mean opinion score) obtained from 21 participants and a quality report describing overall quality, magnitudes, types, and positions of distortions annotated by two participants.}
        \label{fig:mmp2k-example}
        \vspace{-10pt}
\end{figure}

Image quality assessment (IQA)~\cite{SSIM} is a long-standing research topic in image processing. According to the type of assessors, IQA can be divided into subjective IQA and objective IQA, where the former gathers quality ratings from human observers,  and the latter is a quantitative measure that automatically predicts
perceived image quality. 
Although objective IQA is efficient and therefore more suitable for real-life applications, the time-consuming and costly subjective IQA is the only way to collect ground truth ratings, i.e., mean opinion score (MOS), for the development and evaluation of objective IQA metrics. 
Objective IQA can be further categorized into full-reference IQA (FR-IQA), reduced-reference IQA (RR-IQA), and Blind IQA (BIQA) based on the presence of pristine reference image. Among them, Blind IQA is more practical in real-life applications, since reference images are not always available. 

Recently, with the development of computer vision, a number of deep-learning-based BIQA models~\cite{MetaIQA,topiq,hyperiqa,cnniqa,pyiqa} have been proposed and shown promising results. Despite progress, there is limited research on MPIQA. Sang \textit{et al.}~\cite{MP2020} created the MP2020 dataset, consisting of 100 reference MP images and 800 artificially distorted versions with four common distortions, i.e., JPEG, JPEG2000, white noise, and Gaussian blur. However, it is widely believed that the artificial distortions cannot simulate the authentic or in-the-wild distortions in the real world, which are usually a combination of multiple distortions~\cite{KonIQ-10k}. 
Moreover, it is unclear whether the state-of-the-art BIQA models designed for generic natural images can also be effective for MP images with authentic distortions.

To bridge the gap, we conducted a large-scale IQA study on MP images. Specifically, 2,000 MP images were sampled from three public media websites while considering content and quality diversity. We designed a rigorous framework to conduct a subjective study on the sampled images, in which a MOS and a quality report including an overall quality description and distortion magnitudes, types and positions were collected, as shown in Fig.~\ref{fig:mmp2k-example}. The annotations, together with the images, form our multi-labeled MPIQA database, named MMP-2k.

The contributions of this paper are listed as follows.
\begin{itemize}

\item We sourced 15,700 MP images from three public image websites, from which we sampled an image set consisting of 2,000 MP images. This collection of MP images guarantees both content and quality diversity.

\item A lab-based subjective MPQA study was designed and carried out on the image set. 21 participants assessed the quality of each MP image and two participants annotated the magnitudes, types, and positions of distortions in each MP image and generated a quality report\footnote{This work involved human subjects or animals in its research. Approval of all ethical and experimental procedures and protocols was granted by the School of Science and Engineering Research Ethics Committee of the University of Dundee.}.
 
The images, quality ratings, and quality reports form our novel MMP-2k database.

\item We validated the reliability of the MMP-2k database with gold standard questions and self-consistency tests. The experimental results showed that state-of-the-art generic BIQA models underperformed on MP images, indicating the importance of the database for developing quality metrics for MP images.

\end{itemize}

\begin{figure}[htbp]
        \centering
        \includegraphics[width=1.0\linewidth]{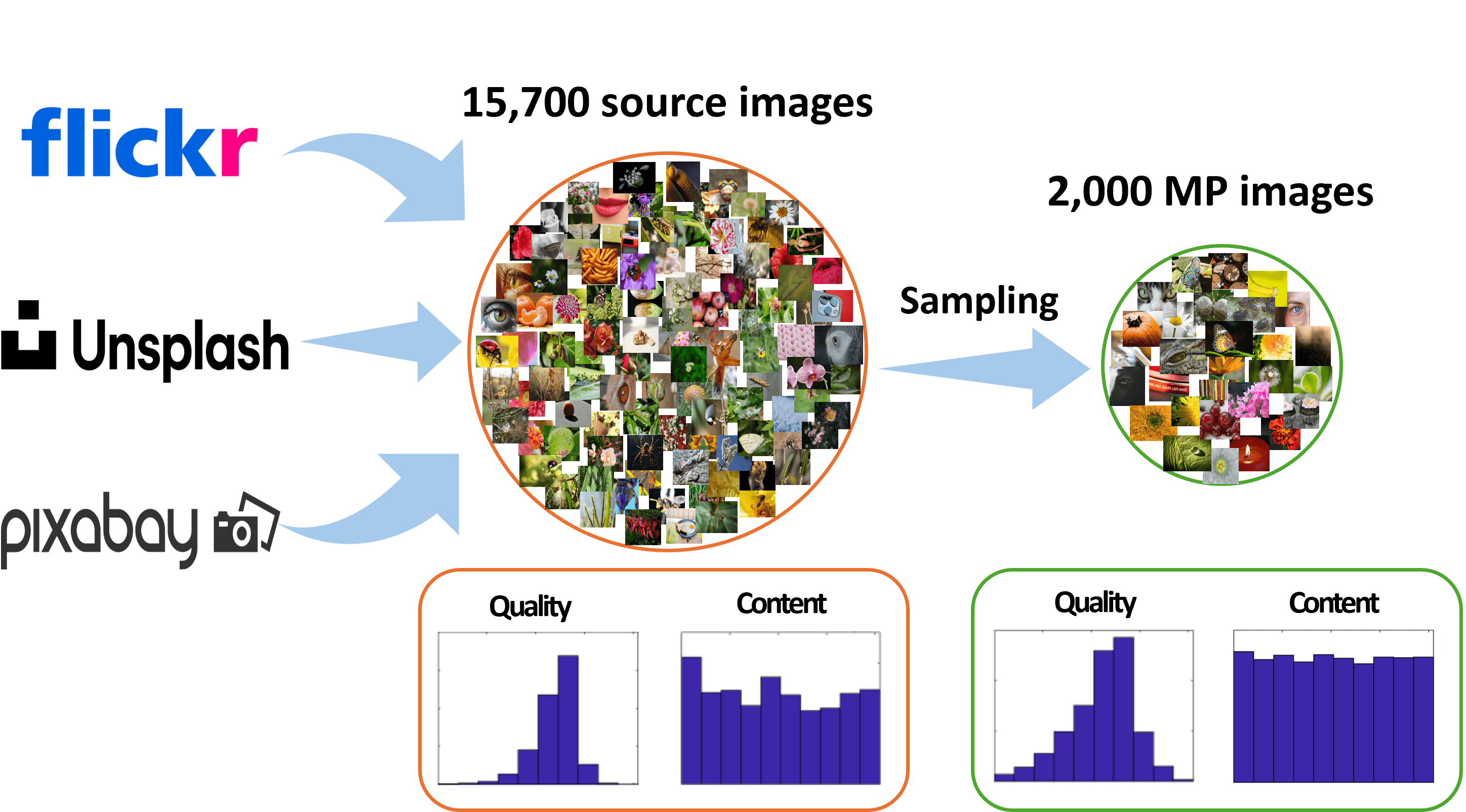}
        \caption{The flowchart of MP image sampling. 15,700 images with a tag of ``macro" or ``macro photography" are acquired from three public image websites, from which 2,000 MP images are sampled considering quality and content diversity.}
        \label{fig:sample_overview}
\vspace{-10pt}
\end{figure}

\begin{figure*}[hbt]
        \centering
        \includegraphics[width=0.95\linewidth]{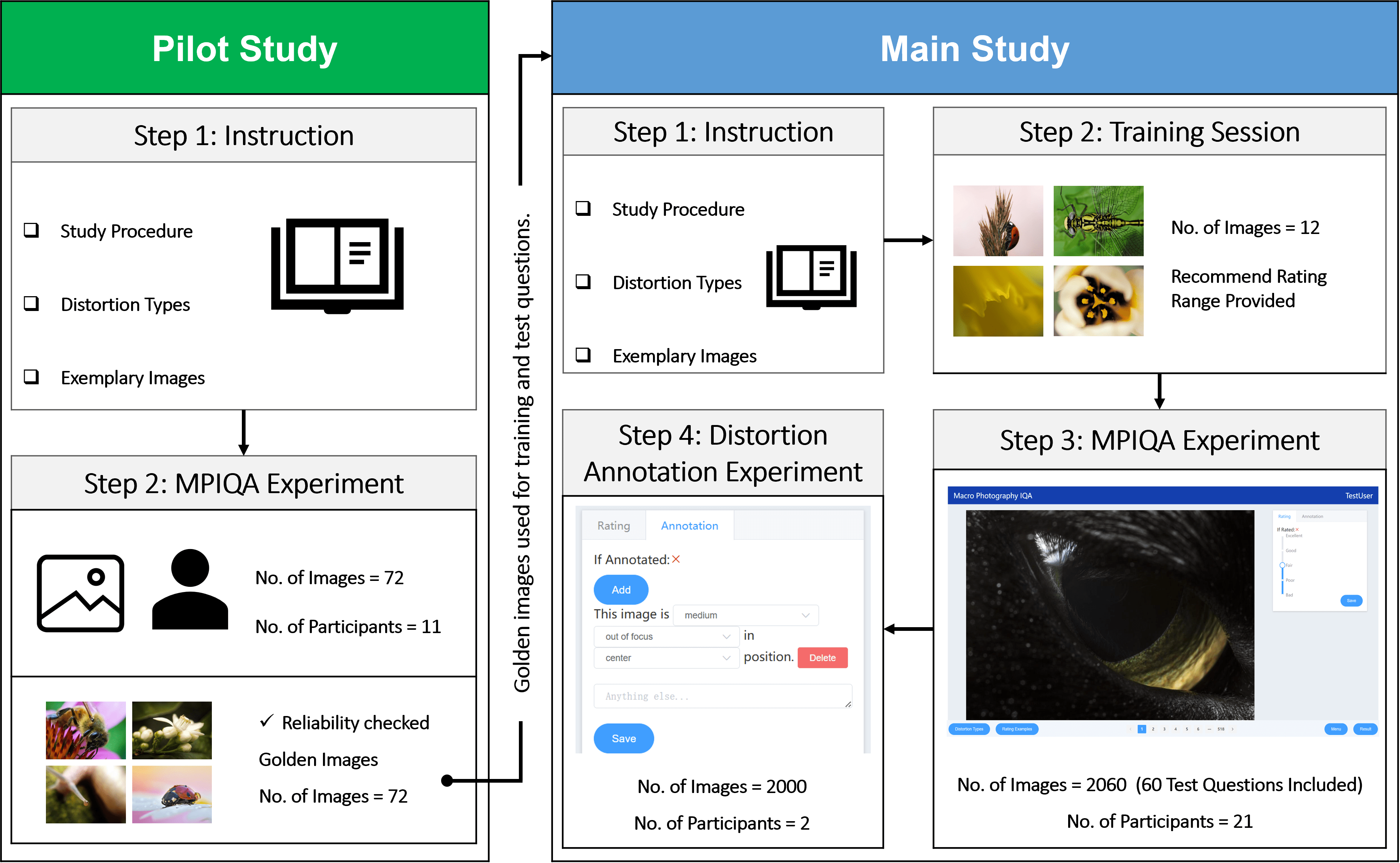}
        \caption{Our subjective MPIQA consists of a pilot study and a main study. In the pilot study, 11 participants were invited to rate the qualities of 72 randomly selected MP images, which were used as training and test questions. In the main study, 21 participants were invited to rate the quality of 2,060 sampled MP images, with training and test questions embedded to enhance and validate participants' reliability. Additionally, two participants annotated the distortion magnitudes, types, and positions.}
      
        \label{fig:workflow}
\vspace{-10pt}
\end{figure*}

\section{MP image sampling}
\label{sec:imageset}

We aim to create an MPIQA database that benefits the development and evaluation of BIQA models on MP images. To this end, collecting MP images with content and quality diversity is important. Fig.~\ref{fig:sample_overview} shows the procedures of MP image sampling, consisting of two stages. In the first stage, 15,700 images with a tag of either ``macro" or ``macro photography" were downloaded from three public image websites. In the second stage, 2,000 MP images are sampled from the downloaded images while ensuring diversity in quality and content.

\subsection{Source images collection}

We collected source images from three public image websites, i.e., \href{https://flickr.com}{Flickr}, \href{https://unsplash.com}{Unsplash}, and \href{https://pixabay.com}{Pixabay}. By using the official application program interface (API) of each website, we requested images with 1) proper common creative licenses permitting commercial use and modifications; and 2) a tag of either ``macro" or ``macro photography". As a result, we downloaded 10,200, 5,822, and 3,000 images, respectively. We manually removed an image if it meets one of the following conditions: 1) it has inappropriate content; 2) it is not an MP image; 3) it has a resolution below 1024$\times$768 (horizontal) or 768$\times$1024 (vertical). For consistency and convenience of later research, all the images were downscaled to 1024$\times$768 (horizontal) or 768$\times$1024 (vertical), maintaining the pixel aspect ratios, followed by central cropping if required. This leads to a set of 15,700 images for further sampling.

\subsection{Diversity sampling}

After looking over the downloaded image set, we found that it has limited content diversity. Specifically, a considerable proportion of MP images captured flowers and insects. Moreover, we noticed that good-quality MP images were predominant, whereas extremely high-quality and low-quality images were deficient. 
One of uses of our MPIQA database is to train better deep learning models. A database with content and quality diversity could improve the generalizability and robustness of models trained and tested on it. Inspired by~\cite{KonIQ-10k}, we used a similar sampling strategy to sample 2,000 MP images while ensuring content and quality diversity. 

To quantify quality, we used the quality score estimated by the Koncept512 model~\cite{KonIQ-10k} as a quality indicator for each image. Although we assume that the BIQA models for generic images may not perform well on MP images, the predicted scores indicate quality diversity to some extent.

To quantify content diversity, we extracted 2048-D deep content features from the pre-trained ResNet-50~\cite{he2016deep} model for each image. The features correspond to the activations of the last `AdaptiveAvgPool2d' layer. Due to each image's high-dimensional deep feature representation, we used $k$-means to quantify them. Specifically, we ran $k$-means ($k$=2050) on the 15,700 image feature vectors and assigned them to the nearest centroid. 
The histogram of the clusters was used for further sampling to ensure content diversity. 

We applied the method proposed by Vonikakis \textit{et al.}~\cite{vonikakis2017probabilistic} for our diversity sampling. It aims to sample a target number of MP images from the large source image set while enforcing a uniform distribution for content and quality indicators, respectively. For the quality indicator, we quantize the predicted scores of Koncept512 into $N$ bins ($N$=2050). After running the sampling algorithm, we finally sampled 2050 MP images from the 15,700 source images, from which we went through the sample images and manually removed 50 images that look like non-MP images.

\section{Subjective MPIQA}
\label{sec:sub_study}



Fig.~\ref{fig:workflow} shows the workflow of our proposed MPIQA, consisting of a pilot study and a main study. The pilot study was conducted to collect participants' feedback to improve the study's graphical user interface (GUI) and generate training and test questions to improve and validate the reliability of participants in the main study. The main study was carried out to assess the qualities of sampled 2000 MP images. In addition, two participants annotated the distortion magnitudes, types, and positions of each MP image and generated a quality report. 

Following the ITU standard~\cite{ITU-BT500}, we applied the 5-point absolute category rating (ACR) scale, i.e., Bad (0), Poor (25), Fair (50), Good (75), and Excellent (100), to assess the quality of MP images. 

\subsection{Pilot Study}

It is important to collect reliable quality ratings. To this end, we conducted a pilot study. On the one hand, we want to test the effectiveness of the designed graphical user interface (GUI) and update it if necessary. On the other hand, we want to collect a few example MP images with ground truth MOSs to train participants and evaluate their reliability. 

\textbf{Step 1: Instruction} Before conducting experiments, the participants were informed that the collected data would be processed according to data protection regulations and signed a consent
form. Next, each participant received instructions, including three sections. The purpose and procedure of the study are introduced in the first section. The common distortions seen in MP images are explained in the second section. In the third section, exemplary MP images with different quality scales are presented. 

\textbf{Step 2: MPIQA Experiment} 72 MP images were randomly sampled from the unsampled source images and used in the pilot study. For consistency, the laboratory environment in the pilot study was set the same as that in the main study. A group of 11 participants participated in the pilot study.

We collected 11 quality ratings for each image. For each image, we calculate the z-score of quality ratings. Ratings with an absolute z-score value greater than 2 were removed as outliers. The remaining ratings were averaged as MOS. By calculating the Pearson linear correlation coefficient (PLCC) between each participant's ratings and MOSs, we found all participants had a PLCC value greater than 0.7, indicating the reliability of ratings. The 72 MP images together with their MOSs are used as \textit{training} and \textit{test} questions in the main study. Based on the feedback from the participants, we also added shortcut keys to speed up participants' rating procedures in the main study.

\subsection{Main Study}

\textbf{Step 1: Instruction} Like the pilot study, participants in the main study signed a consent form and read instructions. 

\textbf{Step 2: Training Session} Participants conducted a training section after reading instructions, in which 12 training questions were provided to guide participants on how to use the interface and rate the quality of MP images. Here, the MP images of 12 training questions were randomly sampled from the 72 MP images in the pilot study. For a training question, the correct rating range of an MP image was set to the interval centered at the ACR scale closest to its MOS and with a width of $\pm$ one ACR scale. 

In the training session, participants click the `Save' button after rating the quality of an MP image. They can proceed to the next training question if their ratings fall into the correct quality range. Otherwise, they are informed that their answer is incorrect, and the correct rating range is suggested. 

\textbf{Step 3: MPIQA Experiment} The sampled 2,000 MP images were divided into four batches and assessed in the main study. To validate the reliability of participants, the remaining 60 MP images evaluated in the pilot study were included as test questions. As a result, there are 515 MP images in each batch, where 15 of them are test questions. 

21 participants were recruited and conducted the MPIQA experiment. During the experiment, participants were allowed to look up the common distortions and exemplary images whenever they wanted. To mitigate the effect of fatigue, the experiment was divided into two sessions, where participants completed two batches in each session. Moreover, participants were required to take a break of 15 minutes after completing the first batch. It took about three to four hours to complete the main study.

\textbf{Step 4: Distortion Annotation Experiment}  The screenshot of user interface for distortion annotation is shown in Step 4 of Fig.~\ref{fig:workflow}. Considering the distinct characteristics of MP images~\cite{adobe2024}, five commonly observed types of distortion in MP images are listed, including `out of focus', `motion blur', `noise and artifact', `overexposure', and `underexposure'.  Three levels, i.e., `mild', `medium', and `strong', are listed to label the magnitude of distortion. Furthermore, 14 options are listed to label the position of distortion, including `entire image', `left', `right', `top', `top left', `top right', `center', `center left', `center right', `center bottom', `center top', `bottom', `bottom left', and `bottom right'. Given the distortion magnitudes, types, and positions, participants were asked to annotate the distortions observed in an MP image. If more than one distortion was observed, participants were allowed to click the button `Add' to annotate another annotation. In addition, participants were allowed to type in a description of observed distortion if the distortion was not listed.

Two participants conducted the distortion annotation experiment. First, each participant annotated observed distortions in each MP image separately. Then, the two participants reviewed their annotations for each MP image, where unanimous annotations were kept directly and disagreeable annotations were decided to be kept or discarded after further discussion to reach a mutual agreement.

\section{Result and Analysis}
\label{sec:result}

\subsection{Outlier Removal and MOS Generation}

\begin{figure}[!htbp]
    \centering
    \begin{subfigure}[b]{0.43\textwidth}
    \centering
        \includegraphics[width=1.0\textwidth]{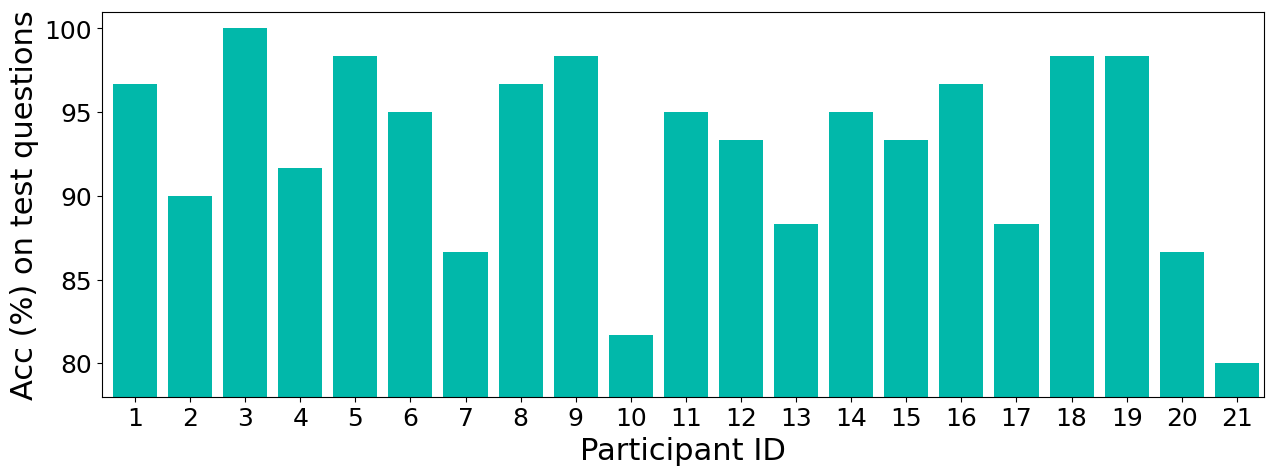}
        \caption{}
    \end{subfigure}
    \begin{subfigure}[b]{0.43\textwidth}
        \centering
         \includegraphics[width=1.0\textwidth]{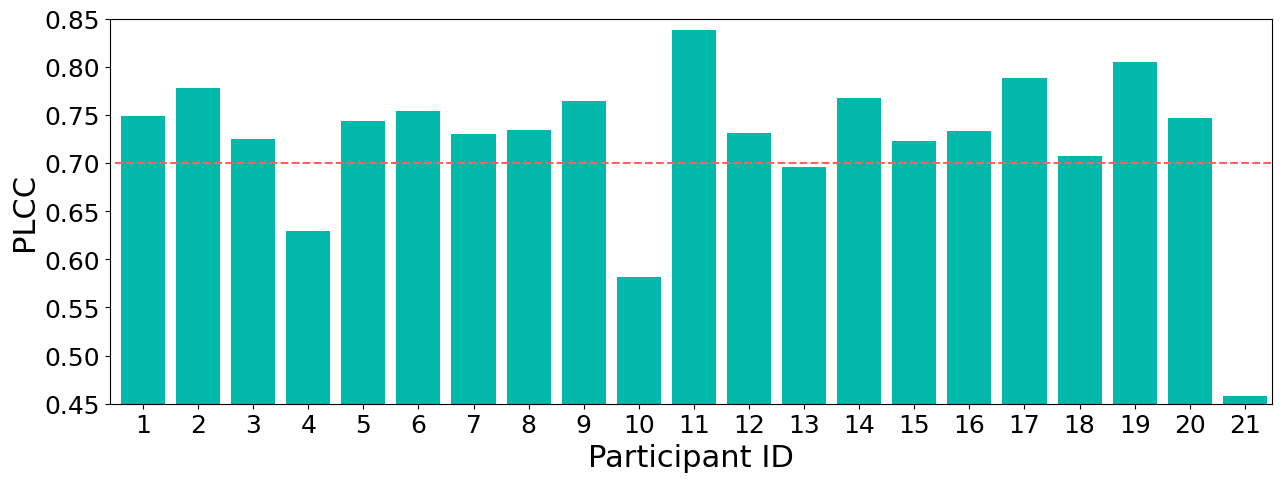}
         \caption{}
     \end{subfigure}
    \caption{Reliability analysis of subjective study. (a) Accuracy of the participants on the 60 test questions. (b)  PLCC correlation between ratings of each participant and MOS. 
    }\label{fig:reliability_analysis}
\vspace{-10pt}
\end{figure}

\begin{figure}[htbp]
        \centering
        \includegraphics[width=0.7\linewidth]{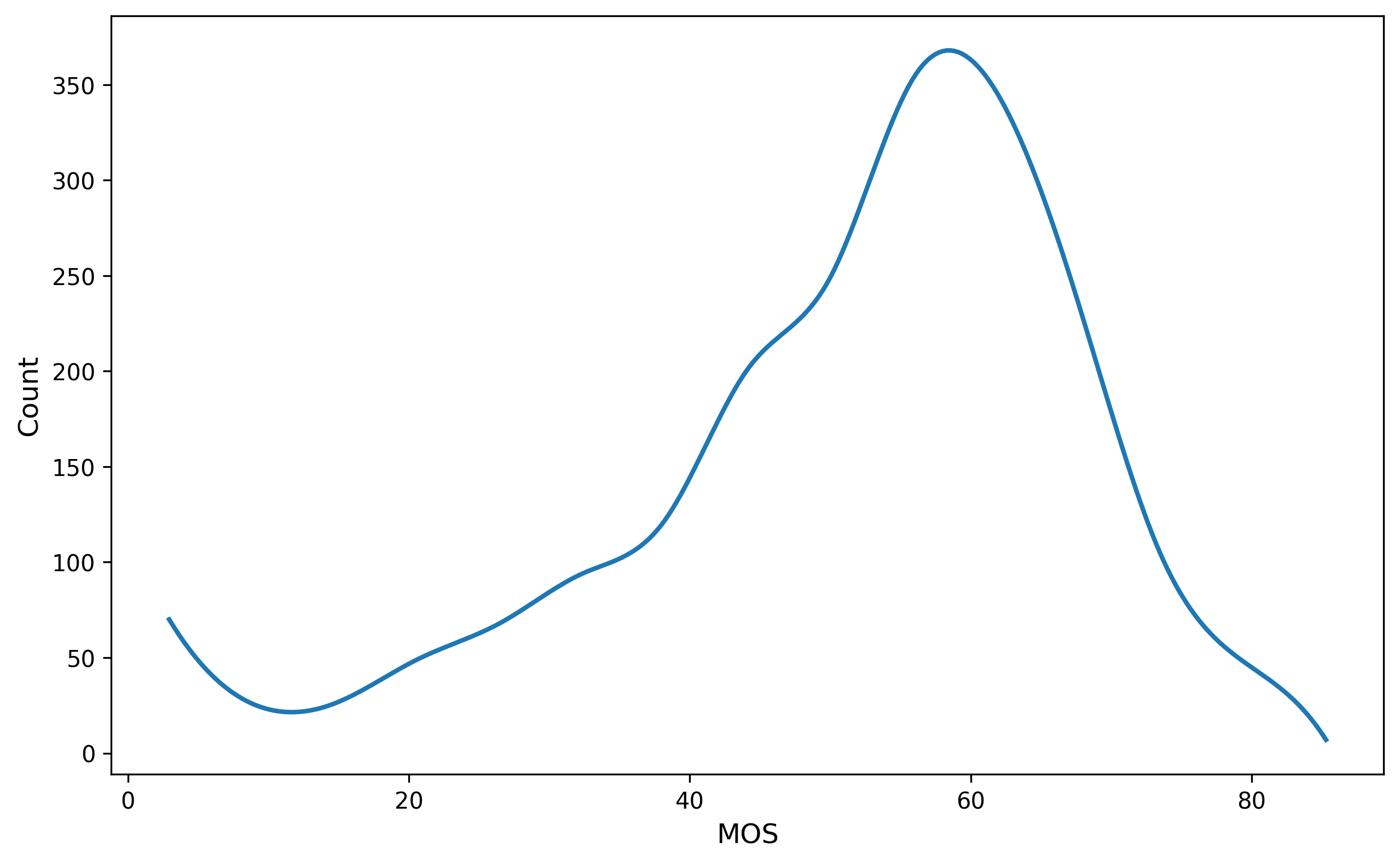}
        \caption{MOS distribution of the MMP-2k database.}
        \label{fig:mos}
\vspace{-10pt}
\end{figure}

We analyze the reliability of participants by measuring their accuracy on test questions and correlation on a self-consistency test, as illustrated in Fig.~\ref{fig:reliability_analysis}. Fig.~\ref{fig:reliability_analysis}(a) reports the accuracy of participants on test questions. As can be seen, all the participants achieve an accuracy of over 80\%, indicating they are eligible for the study. However, by calculating the the Pearson linear correlation coefficient (PLCC) between ratings of each participants and MOS in Fig.~\ref{fig:reliability_analysis}(b), we find that participants 4, 10, 13, and 21 have a PLCC value lower than 0.7. Therefore, we excluded them as outliers, as recommended by \cite{su2023going} and \cite{ITU-BT500}. Eventually, the MOS of an MP image is the average ratings of 17 participants. The MOS distribution of MMP-2k database is shown in Fig.~\ref{fig:mos}.

\subsection{Quality Report Generation}

The quality report of an MP image in the MMP-2k database consists of two parts, including an overall quality description and distortion annotations. 

For the overall quality description, we used a template `The image is $A$ in quality.', where $A$ corresponds to `bad', `poor', `fair', `good', and `excellent' when an image with a MOS of $[0, 12.5)$, $[12.5, 37.5)$, $[37.5, 62.5)$, $[62.5, 87.5]$, and $(87.5, 100]$, respectively. 

For the distortion annotation, we used a template `There is $B$ $C$ in the $D$.', where $B$, $C$, and $D$ correspond to distortion magnitude, type, and position, respectively. When there were more than one distortion annotation, the template was repeated until all of them were listed. Moreover, the description of unlisted distortion was added at the end if it exists.

\subsection{Generic BIQA Evaluation}

As there is a lack of BIQA methods designed for MP images, we selected 11 deep learning-based generic BIQA methods to evaluate on the MMP-2k database, given their strong performance in generic IQA benchmarks LIVE~\cite{SSIM} and KonIQ-10k~\cite{KonIQ-10k}.
The 11 methods are MetaIQA~\cite{MetaIQA}, PI~\cite{pi}, MANIQA~\cite{maniqa}, CLIP-IQA~\cite{clipiqa}, HyperIQA~\cite{hyperiqa}, CNN-IQA~\cite{cnniqa}, DB-CNN~\cite{dbcnn}, ARNIQA~\cite{arniqa}, KonCept512~\cite{KonIQ-10k}, LIQE~\cite{liqe}, and TOPIQ~\cite{topiq}. Except KonCept512, which was implemented by the authors, the other methods are available from IQA-PyTorch~\cite{pyiqa}. For consistency, all input images are downsampled to 512$\times$384 (horizontal) or 384$\times$512 (vertical) before feeding to a model.

Four widely used metrics are applied to evaluate the performance, including PLCC, Spearman rank correlation coefficient (SRCC), Kendall rank correlation coefficient (KRCC), and root mean square error (RMSE). Higher values indicate better performance for all metrics except RMSE, where lower values are preferred.

The performance of state-of-the-art (SOTA) BIQA methods on MMP-2k database is reported in Table~\ref{tb:sota}. Clearly, although the selected BIQA methods have achieved promising performance on generic natural images, they achieve much worse performance on MP images. For example, although the best method, i.e., TOPIQ, achieves 0.984 PLCC and 0.984 SRCC on LIVE database and 0.939 PLCC and 0.926 SRCC on KonIQ-10k database, it merely achieves 0.804 PLCC and 0.762 SRCC on the MMP-2k dataset, which is far from satisfactory. 
It demonstrates that due to the distinct characteristics of MP images, existing SOTA BIQA methods designed for generic natural images are not applicable to MP images. Therefore, there is a need for developing a BIQA method tailored for MP images, and our MMP-2k can serve as a benchmark for its development and evaluation.

\begin{table}[h!]
\centering
\small
\caption{Performance comparison of SOTA BIQA methods on MMP-2k. }
\begin{tabular}{l|cccc}
\toprule
Method & PLCC$\uparrow$ & SRCC$\uparrow$ & KRCC$\uparrow$ & RMSE$\downarrow$ \\
\hline

Meta-IQA~\cite{MetaIQA} & 0.420 
& 0.412 
& 0.286 
& 0.248 
\\
PI~\cite{pi}& 0.598 
& 0.555 
& 0.390 
& 0.246 
\\
MANIQA~\cite{maniqa}& 0.658 
& 0.645 
& 0.468 
& 0.187 
\\
CLIP-IQA~\cite{clipiqa}& 0.679 
& 0.643 
& 0.468 
& 0.158 
\\
HyperIQA~\cite{hyperiqa}& 0.694 
& 0.672 
& 0.492 
& 0.134 
\\
CNN-IQA~\cite{cnniqa}& 0.703 
& 0.641 
& 0.464 
& 0.159 
\\
DB-CNN~\cite{dbcnn}& 0.726 
& 0.728 
& 0.541 
& 0.126 
\\ 
ARNIQA~\cite{arniqa} & 0.739 
& 0.635 
& 0.465 
& 0.141 
\\
KonCept512~\cite{KonIQ-10k}& 0.752 
& 0.653 
& 0.487 
&0.117 
\\
LIQE~\cite{liqe}& 0.755 
& 0.724 
& 0.544 
& 0.226 
\\
TOPIQ~\cite{topiq}& 0.804 
& 0.762 
& 0.580 
& 0.115 
\\ \hline
\end{tabular} \label{tb:sota}
\vspace{-10pt}
\end{table}

\section{Conclusion}
\label{sec:conclusion}
We created MMP-2k, the largest annotated in-the-wild database for macro photography image quality prediction. The database contains 2,000 MP images sampled from a pool of 15,700 images while considering content and quality diversity. Apart from a MOS rating, a quality report is generated to describe the distortion magnitudes, types, and positions that are observed in an MP image. Experimental results demonstrate that the state-of-the-art generic IQA metrics exhibit limited performance on MP images. We expect the database will benefit the development of IQA research.

\bibliographystyle{IEEEbib}
\bibliography{strings,refs}

\end{document}